\documentclass{article}

\usepackage{arxiv}

\usepackage[utf8]{inputenc} 
\usepackage[T1]{fontenc}    
\usepackage{hyperref}       
\usepackage{url}            
\usepackage{booktabs}       
\usepackage{amsfonts}       
\usepackage{nicefrac}       
\usepackage{microtype}      
\usepackage{lipsum}		
\usepackage{graphicx}
\usepackage{cite}
\usepackage{doi}
\usepackage{xcolor}
\usepackage{comment}
\usepackage{multicol,multirow}
\usepackage{array}
\usepackage{colortbl}
\graphicspath{{images/},{.}}
\DeclareGraphicsExtensions{.pdf,.jpeg,.png,.jpg}

\title{x-RAGE: eXtended Reality - Action \& Gesture Events Dataset}

\author{\href{https://orcid.org/0000-0001-7380-0816}{\includegraphics[scale=0.06]{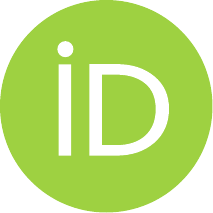}\hspace{1mm}Vivek Parmar$^{1,*}$}, \href{https://orcid.org/0009-0005-3423-8492}{\includegraphics[scale=0.06]{orcid.pdf}\hspace{1mm}Dwijay Bane$^{1,*}$}, 
\href{https://orcid.org/0000-0002-0086-1076}{\includegraphics[scale=0.06]{orcid.pdf}\hspace{1mm}Syed Shakib Sarwar$^2$}, \href{https://orcid.org/0000-0003-3387-054X}{\includegraphics[scale=0.06]{orcid.pdf}\hspace{1mm}Kleber Stangherlin$^2$}, 
\\ \href{https://orcid.org/0000-0002-0810-9903}{\includegraphics[scale=0.06]{orcid.pdf}\hspace{1mm}Barbara De Salvo$^{2,+}$}, \href{https://orcid.org/0000-0003-1417-3570}{\includegraphics[scale=0.06]{orcid.pdf}\hspace{1mm}Manan Suri$^{1,+}$} \thanks{These authors contributed equally.}
        \\
	$^1$ Department of Electrical Engineering, Indian Institute of Technology Delhi, New Delhi, India - 110016 \\
	$^2$ Meta Reality Labs Research \\
 \texttt{barbarads@meta.com, manansuri@ee.iitd.ac.in} \\
}

\renewcommand{\shorttitle}{\textit{x-RAGE}}

\hypersetup{
pdftitle={x-RAGE: eXtended Reality - Action \& Gesture Events Dataset},
pdfsubject={q-bio.NC, q-bio.QM},
pdfauthor={Vivek Parmar, Dwijay Bane, Syed Shakib Sarwar, Kleber Stangherlin, Barbara De Salvo, Manan Suri},
pdfkeywords={Extended Reality, Event Cameras},
}

\begin{document}
\maketitle

\begin{abstract}
With the emergence of the Metaverse and focus on wearable devices in the recent years gesture based human-computer interaction has gained significance. To enable gesture recognition for VR/AR headsets and glasses several datasets focusing on egocentric i.e. first-person view have emerged in recent years. However, standard frame-based vision suffers from limitations in data bandwidth requirements as well as ability to capture fast motions. To overcome these limitation bio-inspired approaches such as event-based cameras present an attractive alternative. In this work, we present the first event-camera based egocentric gesture dataset for enabling neuromorphic, low-power solutions for XR-centric gesture recognition. The dataset has been made available publicly at the following URL: \url{https://gitlab.com/NVM_IITD_Research/xrage}.

\end{abstract}


\section{Introduction}
Bio-inspired event-based sensing that utilizes asynchronous events for efficient data representations presents exciting possibilities to address several limitations concerning memory bandwidth and computational requirements at the edge. Such sensors offer the additional benefit of being able to capture extremely high-speed motions and being unaffected by ambient lighting \cite{9129849} \cite{9138762}. These are important parameters to provide users of XR(extended reality) devices with a truly immersive experience by not constraining environmental factors such as lighting ambience and frame rate of the camera for performing real-time interactions. While XR-centric datasets utilizing ego-centric views have been recently explored in literature\cite{8299578} \cite{8699166} \cite{8088493} there is a gap on usage of event-based sensors for such applications. As part of this study we present the first event-based gesture and actions dataset utilizing the egocentric view for enabling a variety of XR applications \cite{9064606}. This dataset has been collected accounting for effect of ambient lighting, personalization variances in order to present a comprehensive benchmark for enabling real-world deployment.  Following are the salient contributions of this work:
\begin{enumerate}
    \item Egocentric Dataset: First dataset for egocentric gesture and actions utilizing event-based sensors.
    \item Diverse scene dynamics: Incorporates both static and dynamic environments to simulate a range of real-world scenarios, enhancing the dataset's applicability and robustness.
    \item Variable lighting conditions: Features dynamic lighting settings to generate more realistic and challenging event data, better representing the complexities of real-world environments.
    \item Temporal diversity: Includes variations in timing for identical gestures, capturing a spectrum of response times to account for natural human variability in gesture execution.
    \item Comprehensive gesture set: Comprises 36 unique gestures, carefully selected and augmented from previous benchmarks. This set includes 28 single-hand gestures (captured for both hands) and 8 two-handed gestures, offering a rich palette for developing diverse interactive applications.
\end{enumerate}


\section{Experiments}


\subsection{Dataset Collection}
\begin{figure}[t]
\begin{center}
\includegraphics[width=0.6\textwidth]{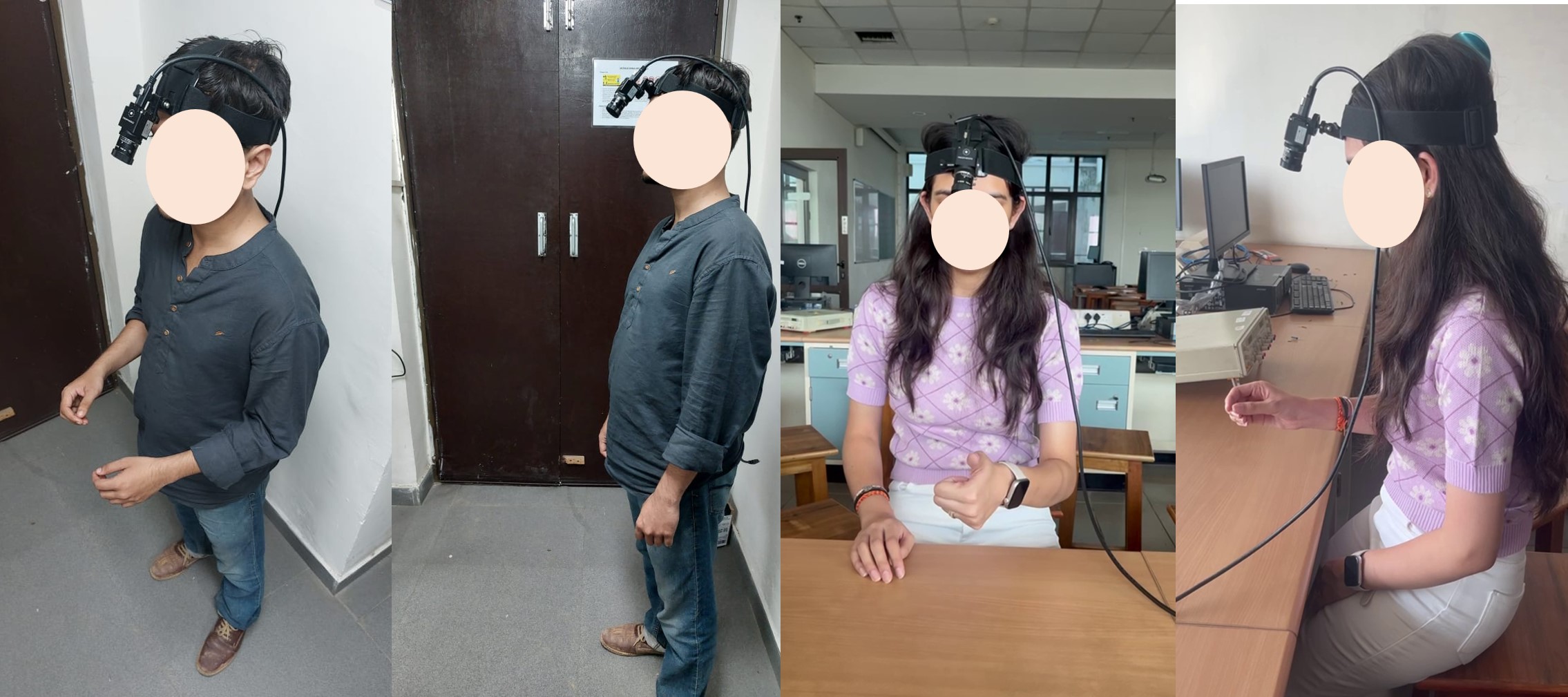}
\end{center}
   \caption{Different perspectives showing the camera setup for data recording for both male and female subjects in standing and sitting positions.}
\label{fig_1}
\end{figure}

\begin{figure}[t]
\begin{center}
\includegraphics[width=\textwidth]{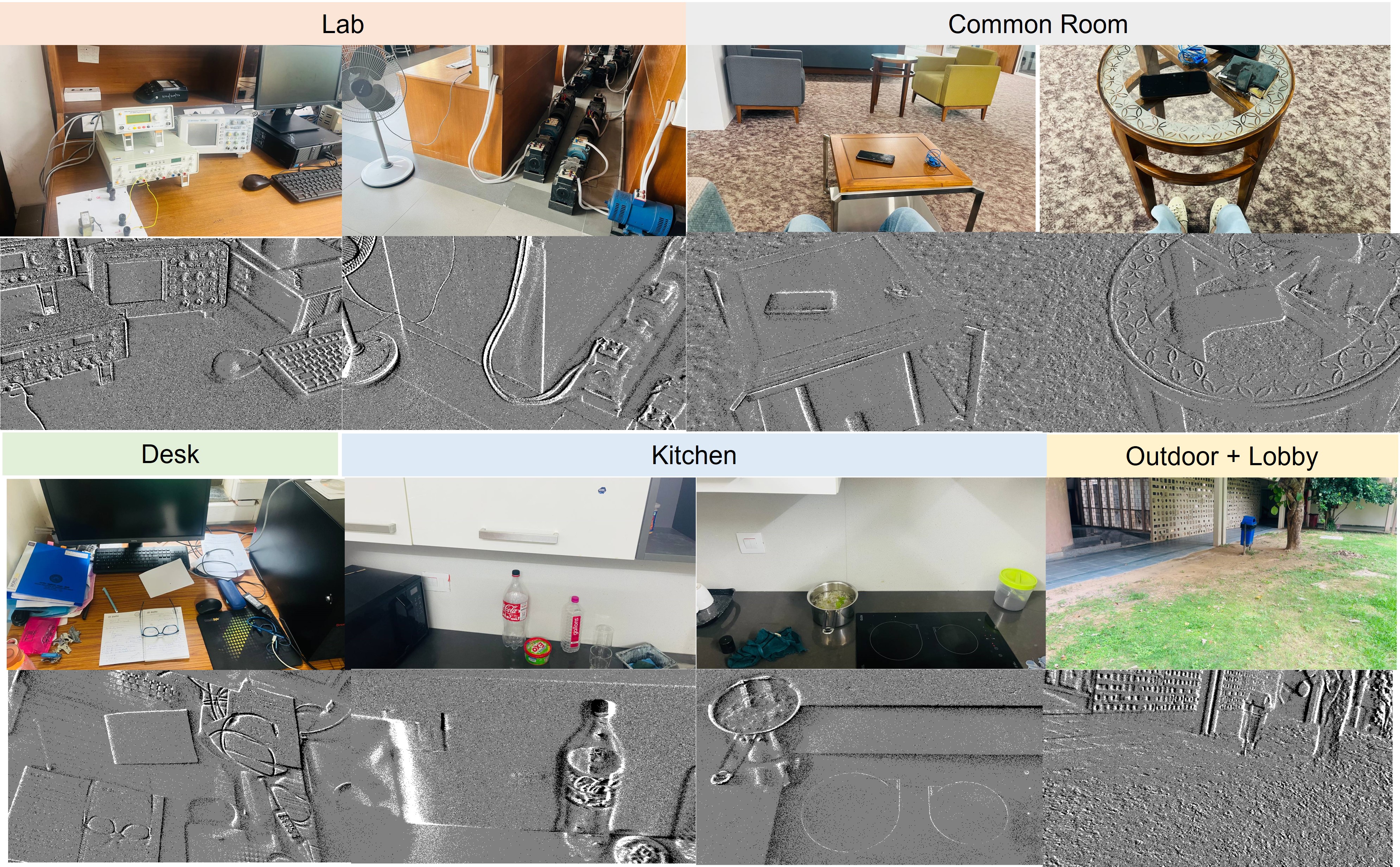}
\end{center}
   \caption{Background scenes used as part of the study shown in both RGB and event data format.}
\label{fig_2}
\end{figure}

The dataset was collected using the Prophesee EVK4 camera. In order to record FPV (first-person view) a special camera setup was developed such that the camera could be strapped on the subject's head (see Fig.~\ref{fig_1}) while connected over USB to a laptop performing the data recording. The camera was tilted to face $\approx$ 60$^\cdot$ downwards. All recordings were performed in 4 indoor and 1 outdoor scenario (see Fig.~\ref{fig_2}). Description of conditions and gestures recorded are provided in Table~\ref{tab1} and Table~\ref{tab2} respectively.

Table~\ref{tab1} maps dynamics of data captured where Condition is scene where the recording is conducted, state is body pose of subject, background mentions if source camera is moving or steady, lighting describes ambient light in scene which possibly needs to be filter if flicker is present or model could be trained with flicker in background, Subjects mentions out of 6 subjects how many did trial for specific condition, samples/class gives count of samples per gesture like for 3 subjects taking 6 samples (\(3 * 6 = 18\)), Total samples denotes Samples/class for all 64 gestures. Actual number of captured samples will be lot higher than total samples as 6 is minimum samples that can be extracted from total samples/class which could be 10-12 samples/class for redundancy. In our dataset we have captured three human poses (sitting, standing and walking), users of wearable devices are frequently in motion, particularly walking. This movement can lead to significant egocentric motion, causing changes in viewing angle and introducing motion blur in typical RGB sensor but in such scenario we can get the benefit of event sensor to capture the gesture.

\subsection{Gesture Classes}
Building upon the EgoGesture benchmark\cite{8299578}, which established a dataset of 83 egocentric gestures using RGB-D sensors, we refined our focus to a subset of 21 gestures. These were selected based on their significant temporal dynamics, excluding static gestures and those easily confused with unintentional movements. Given our use of a 2D event sensor, we also omitted gestures primarily occurring in the depth dimension.

We expanded this selection by introducing 8 new gestures that incorporate pinch expressions and small object rotations, which engage wrist motion. Additionally, we integrated 6 gestures from the NavGesture dataset\cite{maro2020event} for validation purposes.

Our final gesture set comprises 36 unique gestures (see Table~\ref{tab:gesture_table}). Of these, 28 are performed with one hand at a time, and we've captured samples using both hands (yielding 56 gesture variations). The remaining 8 gestures involve simultaneous use of both hands.

This collection draws from Pavlovic's\cite{598226} categorization of hand gestures in human-computer interaction (HCI), encompassing both Manipulative and Communicative gestures. The latter category further divides into symbolic and act-based gestures, which include Mimetic and Deictic subtypes.

\begin{table}[tb]
\caption{Description of 36 unique gestures in categorized form with their variation class number.}
\resizebox{\textwidth}{!}{
\begin{tabular}{|ll|l|l|c|}
\hline
\multicolumn{2}{|c|}{\textbf{Action Category}}                              & \textbf{Gesture}                                     & \textbf{Action}       & \multicolumn{1}{l|}{{\color[HTML]{0D0D0D} \textbf{Available with Both Hands}}} \\ \hline
\multicolumn{1}{|l|}{}              &                                       & pinch drag to top and release                        & Move Top/Forward      & 2, 32                                                                          \\
\multicolumn{1}{|l|}{}              &                                       & pinch drag to right and release                      & Move Right            & 3, 33                                                                          \\
\multicolumn{1}{|l|}{}              & \multicolumn{1}{c|}{Move}             & pinch drag to bottom and release                     & Move Bottom/Backward  & 4, 34                                                                          \\
\multicolumn{1}{|l|}{}              &                                       & pinch drag to left and release                       & Move Left             & 5, 35                                                                          \\
\multicolumn{1}{|l|}{}              &                                       & Fist drag to top and release                         & Move Top/Forward      & 7, 37                                                                          \\
\multicolumn{1}{|l|}{}              &                                       & Fist drag to right and release                       & Move Right            & 8, 38                                                                          \\
\multicolumn{1}{|l|}{}              &                                       & Fist drag to bottom and release                      & Move Bottom/Backward  & 9, 39                                                                          \\
\multicolumn{1}{|l|}{}              &                                       & Fist drag to left and release                        & Move Left             & 10, 40                                                                         \\ \cline{2-5} 
\multicolumn{1}{|l|}{}              & \multicolumn{1}{c|}{Open/Close}       & index pinch \& open palm                             & Select                & 1, 31                                                                          \\
\multicolumn{1}{|l|}{}              &                                       & relax palm to fist                                   & Select                & 6, 36                                                                          \\ \cline{2-5} 
\multicolumn{1}{|l|}{}              &                                       & Zoom in Pinch opening with thumb and index           & Zoom in               & 18, 45                                                                         \\
\multicolumn{1}{|l|}{Manipulative}  & \multicolumn{1}{c|}{Zoom}             & Zoom out Pinch Closing with thumb and index          & Zoom out              & 19, 46                                                                         \\
\multicolumn{1}{|l|}{}              &                                       & left and right hand fist going apart                 & Zoom in               & \cellcolor[HTML]{FFFFFF}20                                                     \\
\multicolumn{1}{|l|}{}              &                                       & left and right hand fist coming close                & Zoom out              & \cellcolor[HTML]{FFFFFF}21                                                     \\ \cline{2-5} 
\multicolumn{1}{|l|}{}              &                                       & Rotating right on a pinch knob with thumb and index  & Clockwise Rotate      & 25, 47                                                                         \\
\multicolumn{1}{|l|}{}              &                                       & Rotating left on a pinch knob with thumb and index   & Anti-Clockwise Rotate & 26, 48                                                                         \\
\multicolumn{1}{|l|}{}              & \multicolumn{1}{c|}{Rotate}           & Rotating right on a bottle cap with thumb and index  & Clockwise Rotate      & 27, 49                                                                         \\
\multicolumn{1}{|l|}{}              &                                       & Rotating left on a bottle cap with thumb and index   & Anti-Clockwise Rotate & 28, 50                                                                         \\
\multicolumn{1}{|l|}{}              &                                       & Rotating right on a small screw with thumb and index & Clockwise Rotate      & 29, 51                                                                         \\
\multicolumn{1}{|l|}{}              &                                       & Rotating left on a small screw with thumb and index  & Anti-Clockwise Rotate & 30, 52                                                                         \\ \hline
\multicolumn{1}{|l|}{}              &                                       & Air Draw Tick                                        & Correct / Accept      & 11, 41                                                                         \\
\multicolumn{1}{|l|}{}              &                                       & Two index finger in cross overlap                    & Incorrect / Decline   & 12                                                                             \\
\multicolumn{1}{|l|}{Communicative} & \multicolumn{1}{c|}{Accept / Decline} & Thumbs Up                                            & Like / Accept         & 13, 42                                                                         \\
\multicolumn{1}{|l|}{(Symbolic)}    &                                       & Thumbs Down                                          & Dislike / Decline     & 14, 43                                                                         \\
\multicolumn{1}{|l|}{}              &                                       & Okay gesture                                         & Okay / Accept         & 15, 44                                                                         \\ \cline{2-5} 
\multicolumn{1}{|l|}{}              & \multicolumn{1}{c|}{Halt / Stop}      & Timeout gesture                                      & Pause                 & \cellcolor[HTML]{FFFFFF}16                                                     \\
\multicolumn{1}{|l|}{}              &                                       & Two hands palm open up                               & Pause                 & \cellcolor[HTML]{FFFFFF}17                                                     \\ \hline
\multicolumn{1}{|l|}{Communicative} & \multicolumn{1}{c|}{Steer Rotate}     & Steer right with left and right fist                 & Clockwise Rotate      & \cellcolor[HTML]{FFFFFF}22                                                     \\
\multicolumn{1}{|l|}{(Mimetic)}     &                                       & Steer left with left and right fist                  & Anti-Clockwise Rotate & \cellcolor[HTML]{FFFFFF}23                                                     \\ \cline{2-5} 
\multicolumn{1}{|l|}{}              & \multicolumn{1}{c|}{Take a Photo}     & Making photo frame with both hands                   & Screenshot            & \cellcolor[HTML]{FFFFFF}24                                                     \\ \hline
                                    &                                       & {\color[HTML]{0D0D0D} Sweeping left}                 & Move Left                   & 53, 55                                                                         \\
                                    &                                       & {\color[HTML]{0D0D0D} Sweeping right}                & Move Right                  & 54, 56                                                                         \\
\multicolumn{2}{|l|}{Navigational Gesture}                                  & {\color[HTML]{0D0D0D} Sweeping up}                   & Move Top/Forward            & 57, 59                                                                         \\
\multicolumn{2}{|l|}{based on NavGesture \cite{maro2020event}}                                   & {\color[HTML]{0D0D0D} Sweeping down}                 & Move Bottom/Backward        & 58, 60                                                                         \\
                                    &                                       & {\color[HTML]{0D0D0D} All fingers apex closing}      & Select                      & 61, 62                                                                         \\
                                    &                                       & Hand waving gesture of greeting "Hi"                 & {\color[HTML]{0D0D0D} Home} & 63, 64                                                                         \\ \hline
\end{tabular}}
\label{tab:gesture_table}
\end{table}

\subsection{Statistical analysis}
\begin{figure}[ht]
\begin{center}
\includegraphics[width=\textwidth]{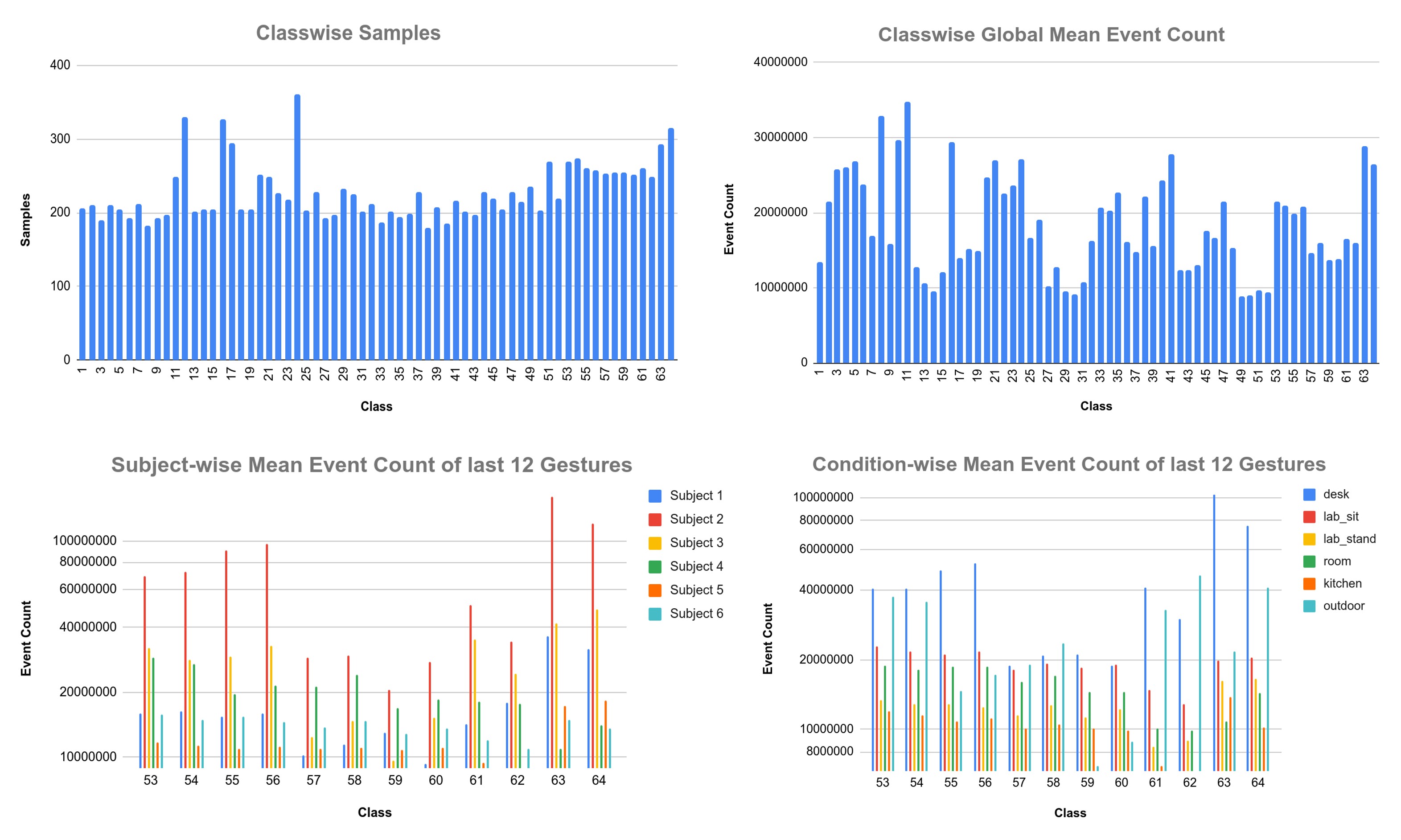}
\end{center}
   \caption{Global statistics: (a) Number of samples per class in the dataset, (b) Mean event counts for each class in the dataset. Statistics limited to NavGesture classes (53-64): (c) Mean Event Count for each subject in each class, (d) Mean Event Count for each scene in each class.}
\label{fig_4}
\end{figure}

\begin{figure}[ht]
\begin{center}
\includegraphics[width=0.8\textwidth]{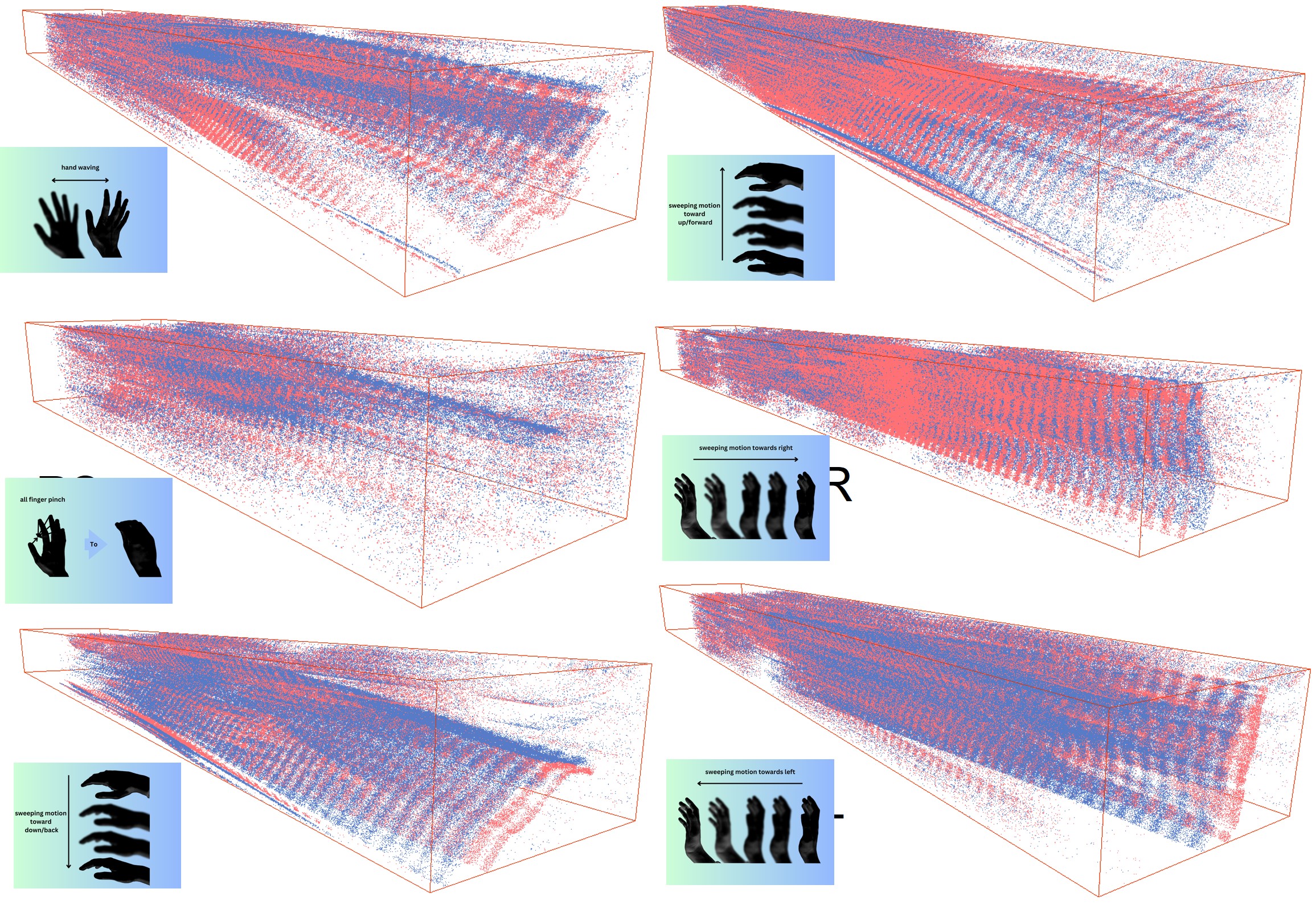}
\end{center}
   \caption{XYT plots showing event surface for 6 classes in Navgesture (right hand).}
\label{fig_5}
\end{figure}

Global statistics concerning number of samples per class and mean event count per class are shown in Fig.~\ref{fig_4}(a) and Fig.~\ref{fig_4}(b) respectively. In order to standardize the benchmark with prior literature 12 gestures based on NavGesture\cite{maro2020event} have been included in the dataset. Impact of subject and environment on the mean event count for each class in NavGesture has been summarized in Fig.~\ref{fig_4}(c) and Fig.~\ref{fig_4}(d) respectively. Event surfaces for a subset of these gestures (i.e. for right-hand) have been shown in Fig.~\ref{fig_5}. 


\begin{figure}[ht]
\begin{center}
\includegraphics[height=18cm]{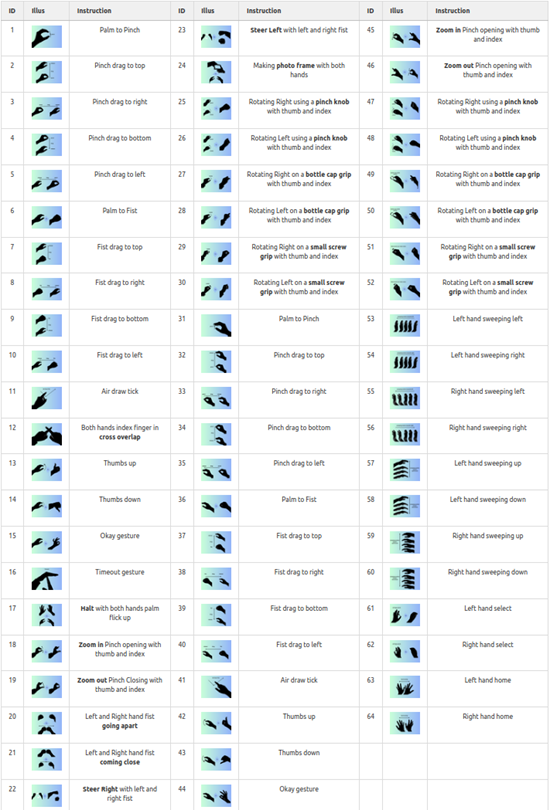}
\end{center}
   \caption{}
\label{fig_all}
\end{figure}

\begin{table}[tb]
  \centering
  \caption{Mapping of complete dataset capturing dynamics of gesture captured and their count.}
  \resizebox{\textwidth}{!}{
    \begin{tabular}{|c|p{5em}|c|l|p{5em}|p{2em}|p{2em}|p{2em}|}
    \hline
    \#    & Condition & State & Background & Lighting & \multicolumn{1}{l|}{\# Subjects} & \multicolumn{1}{l|}{\# Samples/class} & \multicolumn{1}{l|}{\# Total Samples} \\
    \hline
    1     & Indoor Work Desk & \multirow{4}[8]{*}{Sitting} & \multicolumn{1}{c|}{\multirow{5}[10]{*}{Static}} & LED   & 3     & 18    & 1152 \\
\cline{1-2}\cline{5-8}    2     & Lab &       &       & \multicolumn{1}{c|}{\multirow{2}[4]{*}{LED + sunlight}} & 4     & 24    & 1536 \\
\cline{1-1}\cline{6-8}    3     &   Condition    &       &       &       & 4     & 24    & 1536 \\
\cline{1-2}\cline{5-8}    4     & Common Room with Sofa &       &       & Lamp & 4     & 24    & 1536 \\
\cline{1-3}\cline{5-8}    5     & Kitchen & Standing &       & LED   & 4     & 24    & 1536 \\
    \hline
    6     & Outdoor Grass \& Lobby & Walking & Dynamic & Strong Sunlight  & 2     & 12    & 768 \\
    \hline
    \end{tabular}%
    }
  \label{tab1}%
\end{table}%

\subsection{Comparison with literature dataset}
Comparison of the current dataset with event-camera based gesture datasets from prior literature is summarized in Table~\ref{tab2}. The dataset presented in the current study provides the following unique attributes: (i) First-person view, (ii) HD resolution, (iii) Max. number of gestures+actions, (iv) Background variations covering both lighting and background scenes. This effectively leads to a very strong benchmarking capability for future XR devices that can exploit event-based cameras. 

\begin{table}[htbp]
  \centering
  \caption{Comparison with event-based gesture datasets from prior literature.}
    \begin{tabular}{|l|c|c|c|c|c|c|c|}
    \hline
    Dataset & View & Camera & Resolution & \#Classes & \#Subjects & Conditions & \#Samples \\
    \hline
    DVSGesture\cite{dvsg} & SPV   & DVS128 & 128x128 & 11    & 29    & 3     & 1342 \\
    \hline
    NavGesture\cite{maro2020event} & SPV   & ATIS  & 304x240 & 6     & 35    & 2     & 1621 \\
    \hline
    This work & FPV   & Prophesee EVK4 & 1280x720 & 36    & 6     & 6     & 8064 \\
    \hline
    \end{tabular}%
  \label{tab2}%
\end{table}%

\section{Applications}
\label{sec:others}
The dataset is primarily geared towards gesture and action recognition for XR devices. The current release provides time-stamp annotation of the classes. Some possible use cases for this dataset would be:
\begin{itemize}
\item Extended Reality (XR) Interfaces: This dataset is invaluable for developing more intuitive and responsive gesture-based interfaces for extended reality (XR) environments, encompassing virtual, augmented, and mixed reality technologies. The diverse range of gestures and capturing conditions makes it particularly suitable for creating robust gesture recognition systems that can operate in various real-world scenarios.
\item Wearable Technology: The FPV nature of the dataset makes it especially relevant for developing gesture recognition systems for smart glasses, head-mounted displays, and other wearable devices. These applications can benefit from the low latency and high dynamic range properties of event-based vision.
\item Assistive Technologies: Gesture recognition systems developed using this dataset could aid in creating more accessible interfaces for individuals with mobility impairments or in situations where touch-based or voice-based interactions are impractical.
\item Human-Robot Interaction: The dataset can contribute to the development of more natural and intuitive ways for humans to interact with robots, particularly in dynamic environments where traditional vision systems might struggle.
\item Smart Home and IoT Devices: Gesture-based control systems for smart home appliances and Internet of Things (IoT) devices could be developed and refined using this dataset, offering users more intuitive ways to interact with their environment.
\item Automotive Interfaces: The dataset's inclusion of walking scenarios makes it relevant for developing gesture-based interfaces for automotive applications, where both the user and the camera may be in motion.
\item Sign Language Recognition: While not specifically designed for this purpose, the dataset could contribute to the development of more robust sign language recognition systems, particularly in real-world, dynamic environments.
\end{itemize}

To this end, future release of the dataset will incorporate bounding-box data for performing hand tracking.

\section{Conclusions}
\par This study presents a novel First Person View (FPV) event-based actions and gesture recognition dataset, addressing a critical gap in the field of computer vision and human-computer interaction. By utilizing the Prophesee EVK4 event camera, we have created a comprehensive dataset that captures the unique advantages of event-based vision, including high temporal resolution, wide dynamic range, and low latency.
\newline The dataset's strength lies in its diversity and comprehensiveness. With 36 unique gestures per subject, each performed 6 times by 6 different subjects, it provides a robust benchmark for developing and evaluating gesture recognition algorithms. The inclusion of various scenarios - including both stationary and walking conditions - and different lighting environments ensures that the dataset reflects real-world challenges in gesture recognition.
\newline This dataset opens up new possibilities for research and development in event-based vision and gesture recognition. It is particularly valuable for applications in extended reality (XR) and wearable technology, where traditional frame-based approaches may fall short. The dataset's FPV nature makes it especially relevant for developing next-generation interfaces for AR and VR systems. Furthermore, by providing this dataset to the research community, we aim to accelerate the development of more efficient, responsive, and adaptable gesture recognition systems. 

\section*{Acknowledgements}
The authors would like to acknowledge the support of Meta Reality Labs Research and CYRAN AI Solutions.

\bibliographystyle{abbrv}
\bibliography{references}  






\end{document}